\documentclass{article} 
\usepackage{nips15submit_e,times}
\usepackage{hyperref}
\usepackage{url}
\usepackage{booktabs} 
\usepackage{mahnigCEC}

\title{Reproducing and learning new algebraic operations on word embeddings using genetic programming}
\author{
  Roberto Santana \\
  Intelligent Systems Group (ISG) \\
Department of Computer Science and Artificial Intelligence\\
University of the Basque Country (UPV/EHU)\\
San Sebastian, Spain \\
\texttt{roberto.santana@ehu.es} \\
}

\nipsfinalcopy 

\begin{document}

\maketitle

\begin{abstract}
  Word-vector representations associate a high dimensional real-vector to every word from a corpus. Recently, neural-network based methods have been proposed for learning this representation from large corpora. This type of word-to-vector embedding is able to keep, in the learned vector space, some of the syntactic and semantic relationships present in the original word corpus. This, in turn, serves to address different types of language classification tasks by doing algebraic operations defined on the vectors. The general practice is to assume that the semantic relationships between the words can be  inferred by the application of a-priori specified algebraic operations. Our general goal in this paper is to show that it is possible to learn methods for word composition in semantic spaces. Instead of expressing the compositional method as an algebraic operation, we will encode it as a program, which can be linear, nonlinear, or involve more intricate expressions. More remarkably, this program will be evolved from a set of initial random programs by means of genetic programming (GP).  We show that our method is able to reproduce the same behavior as human-designed algebraic operators. Using a word analogy task as benchmark, we also show that GP-generated programs are able to obtain accuracy values above those produced by the commonly used human-designed rule for algebraic manipulation of word vectors. Finally, we show the robustness of our approach by executing the evolved programs on the \texttt{word2vec} GoogleNews vectors, learned over 3 billion running words, and assessing their accuracy in the same word analogy task.  \\
  
{\bf{keywords}}: semantic spaces, compositional methods, \texttt{word2vec}, genetic programming, word vectors
\end{abstract}

\section{Introduction}
  
 In semantic vector word spaces, each word of a given corpus is represented by a vector of real values.  One reason that makes this type of representation relevant is that several natural language processing (NLP) tasks can be efficiently implemented on it. In particular, machine learning methods that use this representation have been proposed for named entity recognition \cite{Turian_et_al:2010}, question answering \cite{Iyyer_et_al:2014}, machine translation \cite{Mikolov_et_al:2013b}, etc.

 Another convenient feature of vector word spaces is that the word vectors are able to capture attributional similarities \cite{Turney:2006} between words. This means that words that appear in similar contexts in the corpus will be close in their vector representation.

 From a machine learning point of view, a crucial question is how meaning can be extracted from the relationships between the vectors. Recent works show that vector word representations obtained using neural networks can capture linguistic or relational regularities between pair of words. For instance, these regularities can be manifested as constant vector offsets between pairs of words sharing a particular relationship \cite{Mikolov_et_al:2013,Mikolov_et_al:2013a}.  Let us use $\overrightarrow{W}$ to represent the vector representation of the word W, then this offset property can be illustrated as  $\overrightarrow{geckos}-\overrightarrow{gecko} \approx \overrightarrow{ants} - \overrightarrow{ant}$. In another example, in the vector space constructed by Mikolov et al,  the algebraic operation $\overrightarrow{king}-\overrightarrow{man}+\overrightarrow{woman}$ will produce a real-value vector whose closest word in the vector word space is ``queen''. More notably, other semantic relationships such as gender-inflections, geographical relationships, etc. can be recovered using algebraic operations between vectors. 

 It has been suggested that the linguistic regularities that vector representations produced by neural networks exhibit are not a consequence of the embedding process itself, but are well preserved by it  however \cite{Levy_and_Goldberg:2014}. This seems confirmed by the fact that for other types of vector representations linear algebraic operations can also produce meaningful results \cite{Levy_and_Goldberg:2014,Pennington_et_al:2014}.

 While it is evident that simple vector algebraic operations such as those aforementioned, capture some semantics encoded in then vector space, it is not clear whether other types of operations could support more precise semantic relationships or unearth more complex or subtle relationships hidden in the semantic spaces. A possible answer to this question could come from  exploring in an efficient way, the space of possible transformations in the vector space so as to find new ways to construct meaning out of word vectors. In this context, genetic programming (GP) \cite{Koza:1992} arises as a natural candidate. 

 GP is a search method that explores the space of programs looking for the one that  maximizes a given evaluation criterion. Programs can be represented in a variety of ways but a common choice is tree-based representation.  Mathematical expressions can be easily represented using a tree in which the  nodes have associated mathematical operators, and every terminal node has an associated operand. Trees are evaluated in a recursive way. The output of the GP tree is contrasted with the desired target value for the input variables, and from this comparison the quality of the tree program is assessed and a ``fitness'' value is assigned to it. A characteristic feature of GP as a search method is that it is evolutionary, i.e.,  a set of programs (population) is progressively modified (evolved) by the application of random modifications (mutations) and  swapping (crossover) of partial trees in the population. 
 
 This paper proposes the use of GP to find a sequence of word vector operations that captures a semantic relationship implicitly encoded in a set of training examples. This constitutes an automatic way to unveil the algebraic operations that express or support a given semantic relationship.  We frame the general question of finding a suitable transformation of word vectors on the more specific \emph{word analogy task} \cite{Mikolov_et_al:2013a,Pennington_et_al:2014}. This task consists of answering a question such as: ``a is to b  as c is to ?''. A correct answer is the exact word that would fit the analogy.  Given the vector representations of the three known words, the problem  to be solved by GP is to produce a vector whose closest word in the corpus is the one that correctly answers the question. 

 Using this particular problem, we address the following research questions: For embedding representations, can meaningful vector algebraic operations be learned  from training examples? If so, is GP a feasible approach to do it? How does GP score with regard to the linear algebraic relationship commonly exploited on vector representations? Are GP evolved programs transferable across linguistic tasks, vector representations and corpora?

The remainder of the paper is structured as follows: In the next section we introduce a general background to vector-based representation of words.  Section~\ref{sec:GP} gives a brief introduction to GP.   Section~\ref{sec:RELWORK} reviews related work.  In Section~\ref{sec:BENCHMARK}, the benchmark of the \emph{word analogy task} dealt with in the paper is described. Section~\ref{sec:GPALG} introduces the approach for  automatically learning compositional methods using GP.  Experiments to evaluate the accuracy of the evolved programs and their transferability across corpora  are presented in Section~\ref{sec:EXPE}. Section~\ref{sec:CONCLU} presents the conclusions of our paper and discusses future work.

\section{Vector-based representations of words}

  In this section we briefly review some of the foundations on which is our work built. We discuss semantic spaces and the approach that creates word embeddings using shallow neural networks. 

\subsection{Semantic spaces}

In semantic spaces, words are given an associated representation  and a number of semantic properties can be inferred from the relationships between these word representations. In this paper we will assume that words are represented  as vectors of real numbers, all the vectors with the same dimension. We will alternatively use the terms ``word vectors''  or ``embeddings'' to refer to the mapping between words and vectors.

 To organize our analysis, we consider two  key issues in semantic spaces:  i) The possible compositional relationships between the word vectors. ii) The methods used to learn the representations. 

  Compositional models are conceived to capture the semantic of a multi-word construction from the semantics of its constituents. The underlying idea is that the vector representations of two or more words could be transformed to obtain the representation of the multi-word construction they form. Algebraic combination of word vectors are of interest in the context of compositional semantics and also relevant for a variety of machine learning tasks in NLP.

To formalize the analysis of methods for word composition, Mitchell  and  Lapata \cite{Mitchell_and_Lapata:2010} define  $p = f(u,v,R,K)$ as the composition of vectors $u$ and $v$. $p$  represents how the pair of words represented by the vectors stand in some syntactic relation $R$, given some background knowledge $K$.

 Compositional methods then propose multiple ways of defining the function $p$. For instance, it is usually assumed that $p$ is a linear function of the Cartesian product of $u$ and $v$, simply defined as
\begin{equation}
  p = Au + Bv  \label{eq:ADDMODEL}
\end{equation}  
where $A$ and $B$ are matrices which determine the contributions made by $u$ and $v$ to $p$.

In \cite{Mitchell_and_Lapata:2010}, additive models such as the one represented by Eq.~\eqref{eq:ADDMODEL} and also multiplicative models are discussed. Other compositional models that consider contextual information (neighboring words)  have been also proposed  \cite{Kintsch:2001}.

Additive and multiplicative compositional models are limited because, among other reasons, they are based on commutative operators that do not attribute any role to the order of the constituents in multi-word constructions. The repertoire of available operations is also constrained if we compare it to the vast range of possible vector manipulations that could be encoded by a more general ``program''. 

 The second issue relevant for semantic spaces is how are they created. Vector spaces can be learned from computing statistical measures of correlations between words or learned by capturing word-context relationships while scanning the sentences of a given corpus.  Neural networks have been applied for implementing the latter approach \cite{Collobert_and_Weston:2008,Zhila_et_al:2013}.   

  In this paper we have used vectors learned by the application of shallow neural networks as proposed in \cite{Mikolov_et_al:2013a}. In the following section we briefly review this approach.

\subsection{Learning word embeddings using neural networks}

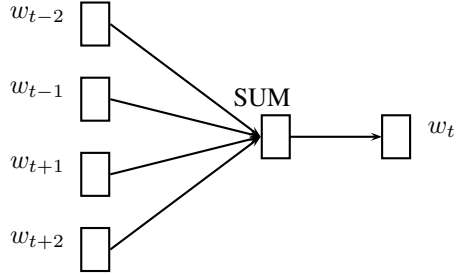
\begin{figure}  
\begin{center}  
\begin{pspicture}(-1,0)(4,4)  

\rput(0,4.0){\rnode{W1}{\psframe(-0.4,-0.3)(0.0,0.3)\rput[br]{N}(-0.6,0){$w_{t-2}$}}}
\rput(0,3.0){\rnode{W2}{\psframe(-0.4,-0.3)(0.0,0.3)\rput[br]{N}(-0.6,0){$w_{t-1}$}}}
\rput(0,2.0){\rnode{W3}{\psframe(-0.4,-0.3)(0.0,0.3)\rput[br]{N}(-0.6,0){$w_{t+1}$}}}
\rput(0,1.0){\rnode{W4}{\psframe(-0.4,-0.3)(0.0,0.3)\rput[br]{N}(-0.6,0){$w_{t+2}$}}}

\rput(2.0,2.5){\rnode{S}{\psframe(0.0,-0.3)(0.4,0.3)\rput[br]{N}(0.4,0.4){SUM}}}
\rput(4.0,2.5){\rnode{WT}{\psframe(-0.4,-0.3)(0.0,0.3)\rput[br]{N}(0.6,0){$w_t$}}}

\ncline{->}{W1}{S} \ncline{->}{W2}{S}
\ncline{->}{W3}{S} \ncline{->}{W4}{S}
\ncline[nodesep=11pt]{->}{S}{WT} 
\end{pspicture} 
\caption{Continuous Bag-of-Words (CBOW) model as proposed in \cite{Mikolov_et_al:2013a}.}  
\label{fig:CBOW}  
\end{center}  
\end{figure}

 In \cite{Mikolov_et_al:2013a}, two neural-network based models have been proposed to learn embeddings: Skip-gram and Continuous Bags of words (CBOW) models. Skip-gram learns to  predict the surrounding words of a given word in a sentence. CBOW learns to predict, given the surrounding words, the word most likely to be in the center.  We focus on the CBOW model.

 CBOW is a feed-forward neural net language model \cite{Bengio_et_al:2003} with a number of added changes. The most important difference is that the hidden layer has been removed. The rationale behind this modification was to explore simpler models. They can not represent the non-linear interactions that neural networks with hidden layers can, but they are much more efficient for learning from millions of words. The CBOW network also uses a Huffman binary tree for more efficient representation of the word vocabulary and a hierarchical softmax scheme. 

 Figure~\ref{fig:CBOW} shows a schematic representation of the CBOW architecture  \cite{Mikolov_et_al:2013a}. Learning is done by scanning the corpus and considering, for each target word $w(t)$, a window comprising words from  $t-k$ to $t+k$ where $2k$ is the window size. In the results reported in  \cite{Mikolov_et_al:2013a}, the best results were obtained using $k=4$. The model was trained using stochastic gradient descent and backpropagation.  

\subsection{Generation of the embeddings}

To generate the embeddings we work with in this paper, we have used the  \texttt{text8.zip} corpus\footnote{Available from \url{http://mattmahoney.net/dc/text8.zip}}. This corpus has been extracted from the English Wikipedia\footnote{Details on the procedure to extract the data are available from \url{https://cs.fit.edu/\%7Emmahoney/compression/textdata.html}}. It comprises $71291$ words. 

 We use the  original \texttt{word2vec} implementation\footnote{\url{http://code.google.com/p/word2vec}} of Mikolov et al \cite{Mikolov_et_al:2013,Mikolov_et_al:2013a} to train the CBOW network from the corpus and generate embeddings. The parameters used by the \texttt{word2vec} program to generate the embedding are described  in Table~\ref{tab:W2V_PARAMS}.

 \begin{table}
 \begin{center}  
  \begin{tabular}{|c||c|c|c|c|}
     \hline
     Parameter & vector size  & Window  &  negative & hs   \\ \hline 
     Value   & 200      & 8       &  25     & 0   \\\hline
     Parameter & sample   & threads &binary     & iter  \\\hline
      value          & 1e-4         & 6       &  1     & 15  \\\hline
  \end{tabular}
  \caption{Parameters used by \texttt{word2vec} to train the CBOW model.}
  \label{tab:W2V_PARAMS}
  \end{center} 
\end{table}

The CBOW is only generated once, regarding to the GP implementation, the most important parameter is the vector size. A larger vector size may allow a more accurate representation of the words. However, the vector size also influences  the computational cost of the algebraic operations between the words that are applied intensively while GP searches for an optimal way to compose the words.

 To evaluate the scalability and robustness of the programs evolved by GP, we also used a much larger embedding. The \texttt{word2vec} word vector model\footnote{Available from \url{https://github.com/mmihaltz/word2vec-GoogleNews-vectors}} comprises 3 million 300-dimension English word vectors and was trained with the Google News corpus (3 billion running words).

\section{Genetic programming} \label{sec:GP}

Genetic programming \cite{Koza:1992, Poli_et_al:2008} is a domain-independent method for the automatic creation of programs that solve a given problem. Each GP program can be seen as a candidate solution to the problem. The process to find the optimal solution is posed as a search in the space of possible programs. The search is organized using a traditional evolutionary optimization approach in which sets (populations) of programs are evolved and transformed by the application of the so-called mutation and crossover operators.

\begin{BAlgo}{GP algorithm}
 \label{alg:GP}
 \item  $D_{0}$ $\leftarrow$ Generate $M$ GP individuals randomly and evaluate them using the fitness function.
 \item $l=1$
 \item \Do
  \item \T {Select a population $D^s_{l}$ from $D_{l-1}$ according to a selection method}
 \item \T {Create a population $D_{l}$ applying genetic crossover to individuals in $D^s_{l}$  with probability $p_{cx}$}
 \item \T {Apply mutation to individuals in $D_{l}$ with probability $p_{mt}$}
 \item \T{Evaluate the individuals in  $D_{l}$}  
  \item \T {$ l \rightarrow l+1$}
 \item  \Until{A stop criterion is met}
\end{BAlgo}

Algorithm~\ref{alg:GP} shows the pseudocode of a very general GP algorithm. Issues in the application of GP are the choice of the program representation, the algebraic operators used by the program, and the objective or fitness function to evaluate the programs. We will discuss these issues in more detail in Section~\ref{sec:GPALG}. However, in order to build some intuition on the particular way in which GP is used in this paper, we present a simple example of the representation.

\begin{figure}[htb]
\begin{center}    
\includegraphics[width=3.0cm]{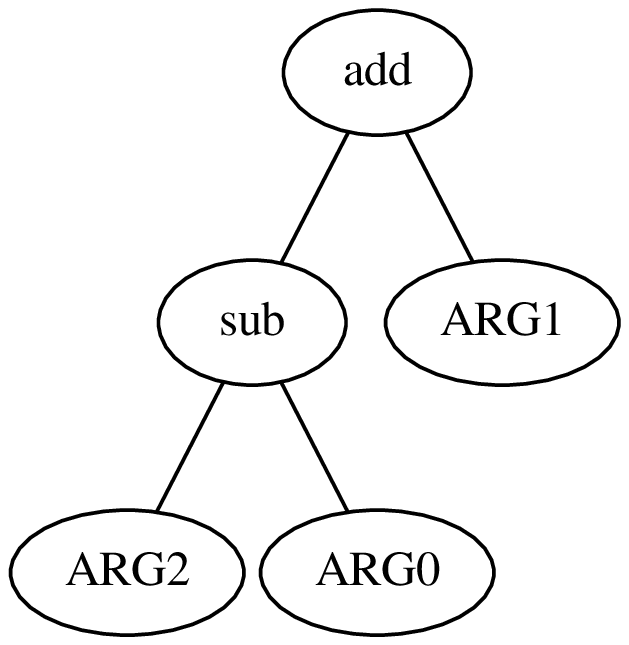}
\includegraphics[width=3.0cm]{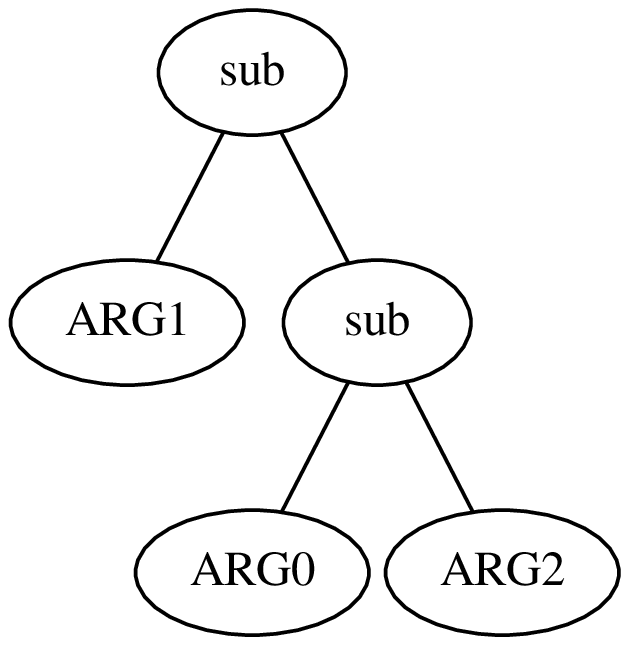}
\includegraphics[width=3.0cm]{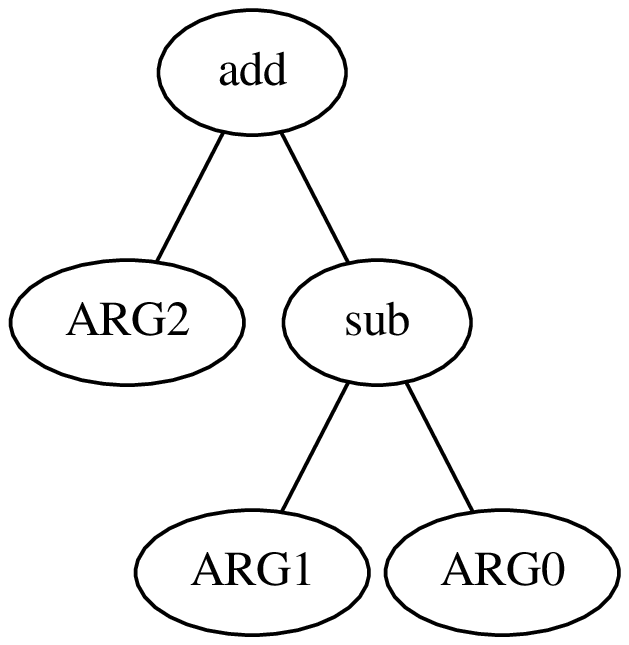}
\includegraphics[width=3.0cm]{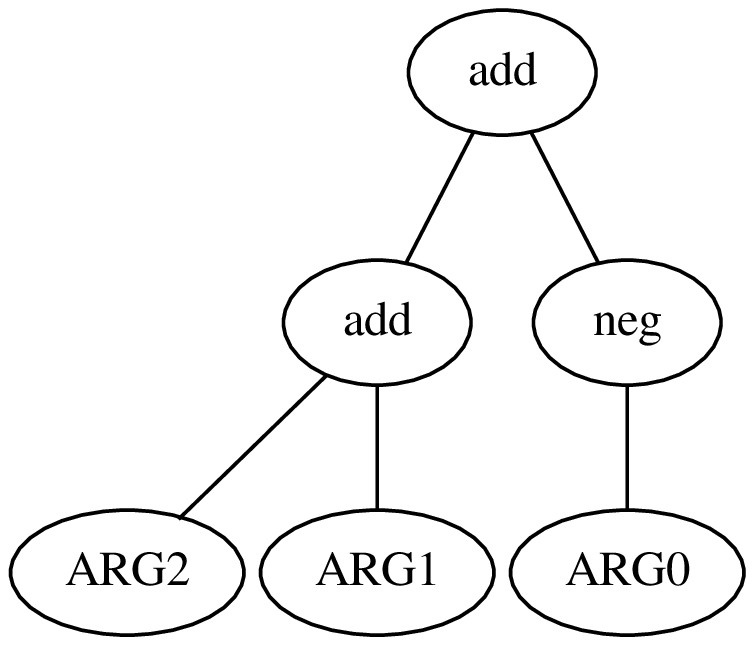}
\caption{Four programs evolved by the GP algorithm. All implement the linear algebraic rule $\vec{d}=\vec{c}-\vec{a}+\vec{b}$.}
\label{fig:EQUIV_TREES}
\end{center}  
\end{figure}

 Let us consider that the three words in the question ``a is to b  as c is to ?'' are transformed to their vector representations, which will be the three arguments of a program. They are transformed as: $a \rightarrow ARG0$, $b \rightarrow ARG1$, $c \rightarrow ARG2$. Then, the linear algebraic rule to compute the answer to the questions, i.e., $\vec{d}=\vec{c}-\vec{a}+\vec{b}$, could be represented as $add(ARG2,sub(ARG1,ARG0))$, where $add$ indicates addition, and $sub$, subtraction. Figure~\ref{fig:EQUIV_TREES} shows four GP programs that produce the same rule. The representation shown in Figure~\ref{fig:EQUIV_TREES} is called a tree-based GP representation and is the one used in this paper. The tree representation is a convenient way to recursively organize the evaluation of a particular composition of the word vectors. Depending on the set of available operators (those defined in the non-terminal nodes of the trees) a richer space of possible word vector compositions could be represented. What the GP algorithm does is to bias the search toward those programs that maximize the given fitness function.

\section{Related work} \label{sec:RELWORK}

Levy and Goldberg \cite{Levy_and_Goldberg:2014} investigate  the question of how to recover the relational similarities in word embeddings. They show that the linear algebraic proposed by Mikolov et al. \cite{Mikolov_et_al:2013a} to solve analogy recovery is equivalent to searching for a word that maximizes a linear combination of three word similarities. Furthermore, they propose an alternative way to compute the distance between the vector generated and the set of words from the corpus. While our research is related to the work presented in \cite{Levy_and_Goldberg:2014}, our focus is on the operations involved in the generation of the candidate word vector, not on the way the match between the generated vector and the word vectors in the corpus is assessed.

Pennington et al  \cite{Pennington_et_al:2014}  introduce global log-bilinear regression models as an alternative to shallow neural-networks to produce word embeddings.  They show their model is able to produce a word vector space with meaningful substructure. The algebraic rule they use to solve the  \emph{word analogy task} is the same as that originally introduced in \cite{Mikolov_et_al:2013a}. Although they applied the distance measure previously presented by Levy and Goldberg \cite{Levy_and_Goldberg:2014}, they report that this distance did not produce better results than the original one. The work presented in \cite{Pennington_et_al:2014} is relevant for our research since it confirms that the usability of the word vector algebraic rule extends over vector representations obtained using a variety of model types and algorithms. 

Socher et al \cite{Socher_et_al:2011,Socher_et_al:2011a} propose a method for compositional representation of words that learns a binary parse tree for each input phrase or sentence. The leaves of the tree contain vector representation of words. The tree serves as the basis for the application of a deep recursive autoencoder. Although this representation uses a tree structure to combine the word vectors, it is completely different to a GP approach. Furthermore, trees are independently inferred for each single sentence or multi-word phrase.

Grefenstette et al \cite{Grefenstette_et_al:2011} propose associating different levels of meaning for words with different types of representations.  For example, verbs or other relational words would be represented by matrices while nouns as vectors. Algebraic operations involving matrices and vectors are used to produce sentence vectors. In principle, GP approaches could cater for joint use of vector and matrix representation by means of strongly typed GP \cite{Montana:1995} or other GP variants that guarantee type constraint enforcement. However, it makes more sense to exhaust the potential of  homogeneous word representations before recurring to GP based on more complex word representations.

In \cite{Blacoe_and_Lapata:2012}, three word vector representations and three compositional methods (addition of word vectors, multiplication of vectors, and the aforementioned deep recursive autoencoder approach) are combined to evaluate their applicability to estimate phrase similarity and  paraphrase detection (i.e., determining whether two sequences have the same meaning). The reported results show that diverse combinations of representations and compositions produce the best results for different problems. Authors state ``the  sizes of the involved training corpora and the generated vectors are not as important as the fit between the meaning representation and compositional method.'' This fact highlights the importance of finding an appropriate compositional method. 

As a summary of this brief review of related work on compositional methods, we point out that although several papers emphasize the important role of these methods for solving an array of semantic tasks,  we did not find any previous report of the automatic learning of the word compositions. 

It is important to notice that from the point of view of machine learning problems, the \emph{word analogy task} is not a classification problem. This is so even if the quality of a solution can be given in terms of accuracy, as the fraction of correctly answered questions. Neither it is  a classical regression problem since each single  input and output feature is represented using a vector of high-dimensional variables. In this context GP has been less investigated than for classical classification and regression problems. However, GP has been applied to a miscellany of tasks in information retrieval \cite{Cummins_and_ORiordan:2006,Escalante_et_al:2015,Oren:2002,Trotman:2005}. In particular, Oren \cite{Oren:2002} combines GP with vector-based representation of documents for information retrieval.  Other problems that involve text classification have also been addressed with GP. Two related areas where GP has been applied are document ranking \cite{Trotman:2005} and term-weighting learning \cite{Cummins_and_ORiordan:2006,Escalante_et_al:2015}. We did not find any previous report on the combination of genetic programming and word embeddings.


\section{Problem benchmark:  \emph{word analogy task}}  \label{sec:BENCHMARK}

The  \emph{word analogy task} consists of answering a question such as: ``a is to b  as c is to ?'' A correct answer  is the  exact word that would fit the analogy. Table~\ref{tab:QUESTION_EXAMPLES} shows several exemplar questions. We used the  benchmark proposed by Mikolov et al. \cite{Mikolov_et_al:2013}  in which questions are separated into $13$ groups. In Table~\ref{tab:QUESTION_EXAMPLES}, Group refers to the group from which the example was taken.

 \begin{table}[tb]
  \centering
  \begin{tabular}{|c|c|c|c|c|}  \hline
    Group & word 1    & word 2       & word 3    &  Answer      \\ \hline \hline
    4 &    boy      &  girl        & sons      & daughters \\ \hline
    5 & amazing     &  amazingly   &   slow    & slowly    \\ \hline
    6 & honest      & dishonest    &  known    & unknown   \\ \hline
    7 &          bad&       worse  &     old   & older     \\ \hline
    8 &          bad&       worst  &    good   & best      \\ \hline
    9 &         code&       coding &    walk   & walking   \\ \hline
    11&      dancing&       danced & feeding   & fed       \\ \hline
    12&       banana&      bananas &     car   & cars      \\ \hline
    13 &    decrease&  decreases   &   say     & says      \\ \hline
  \end{tabular}
    \caption{Examples of questions  in the \emph{word analogy task}.}
  \label{tab:QUESTION_EXAMPLES}

\end{table}

 Table~\ref{tab:QUESTION_TASK} shows the description of the \emph{word analogy task} benchmark. In this table, $Nq_{orig}$ is the number of questions in the original benchmark and $Nq$ is the number of question after removing those words that do not appear in the shortened corpus we used in our experiments. Since the corpus we use is relatively small, for $4$ of the $13$ groups of questions (``capital-world'', ``currency'', ``city-in-state'', ``nationality-adjective'') we did not find one or more of the four words for each of the questions. Therefore, these four groups of questions were excluded from our analysis.

\begin{table}[tb]
  \centering
  \begin{tabular}{|c|c|r|r|}  \hline
      Group & Name  & $Nq_{orig}$ & $Nq$        \\ \hline \hline
      4   & family (gender inflections) &506 & 305     \\ \hline
      5   & gram1-adjective-to-adverb   &992 & 755     \\ \hline
      6   & gram2-opposite               &812 & 305     \\ \hline
      7   & gram3-comparative            &1332 & 1259   \\ \hline 
      8   & gram4-superlative            &1122 & 505    \\ \hline
      9   & gram5-present-participle     &1056 & 991    \\ \hline
      11  & gram7-past-tense             &1560 & 1331   \\ \hline
      12  & gram8-plural (nouns)         &1332 &  991   \\ \hline
      13  & gram9-plural-verbs           & 870 &  649   \\ \hline 
  \end{tabular}
    \caption{Description of the \emph{word analogy task} benchmark, where $Nq_{orig}$ is the number of questions in the original database and $Nq$ is the number of question after removing those words that do not appear in shortened corpus.}
  \label{tab:QUESTION_TASK}
\end{table}

\section{Description of the GP approach} \label{sec:GPALG}

The automatic learning of the composition of words is possible in the specific problem we use, and the GP task, given the vector representations of three words that define a question, is to produce a vector whose closest word in the corpus is one that correctly answers the question. We will mainly use the CBOW model learned using \texttt{word2vec} to determine which is word vector of the model encoding a given word, or to find which is the word in the model whose encoding vector is the closest to a target word vector.  The pseudocode of the GP algorithm we used is shown in Algorithm~\ref{alg:GP}. It is a straightforward implementation of tree-based genetic programming.

 The selection method used is \emph{truncation selection}. After sorting the individuals according to their fitness, the best $100$ solutions are kept for crossover and mutation. \emph{Uniform mutation} randomly selects a point in the tree individual and replaces it by a random subtree. \emph{One-point crossover} is used, it randomly selects two subtrees in the individuals and then exchanges them. The probability of mutation and crossover was $p_{m}=p_{cx}=0.5$.

 The choice of genetic operators has been made as simple as possible to enhance the readability of the algorithm. While more sophisticated GP methods exist, our focus here is the proof of concept of automatic generation of the compositions, and, for that purpose, the choice of the operators was appropriate. On the other hand, we conducted a set of preliminary experiments with other mutation and selection operators\footnote{Those included in the DEAP library used to implement the algorithms} and did not appreciate significant changes in the results when the set of all groups of analogy questions were considered. Some operators can produce more accurate programs for some particular group, but then they  are outperformed by other methods in other groups.   

 In the experiments, the population size used was $N=500$ and the stop criterion  is a maximum number ($ngen=250$) of generations.  Our GP implementation was written in Python. It is based on the EA software \texttt{DEAP}\footnote{\url{http://deap.readthedocs.io/en/master/api/tools.html}} \cite{Fortin_et_al:2012}  and the \texttt{gensim} package\footnote{\url{https://radimrehurek.com/gensim/}}, a Python-based implementation of NLP algorithms \cite{Radim_and_Sojka:2010}. \texttt{gensim} includes methods for interrogating the model generated by  \texttt{word2vect}. Our code is openly available \footnote{\url{https://github.com/rsantana-isg/GP_word2vec}}.

\subsection{GP operators}

 The set of operators used by the programs is shown in Table~\ref{tab:GPParts}. All operators are defined on vectors of the same dimension. There are two classes of operators: binary and unary.  The $+$, $-$ and $*$ operators have the following meaning: vector addition, vector subtraction, and vector component-wise multiplication, respectively; while $\%$ corresponds to \emph{protected} division (usual division except that a divide by zero for any of the vector components returns zero). We discarded the possibility of including  fixed (vector) random constants as terminals of the programs since they may depend on the size of the vector and our aim was to produce programs scalable to any vector dimension. We set a constraint $d=10$ to the depth of the trees to reduce the complexity of the programs.

\begin{table}[tb]
  \centering
  \begin{tabular}{|c|c||c|c||c|c|}
    \hline
  B.   & w $op$ v      & U. & $op$ w& U. & $op$ w  \\         \hline  \hline
       add     & $w+v$         &neg      & $-w$     &Roll &  $(w_2\cdots,w_n,w_1)$    \\\hline
       sub     & $w-v$         &diff     & $1+w$    &Rint &  $int(w)$               \\\hline
       mul     & $w*v$         &abs      & $abs(w)$  &Half & $\frac{1}{2}w$       \\\hline
       saveDiv & $w\%v$         &cos      & $cos(w)$ &Norm &  $\frac{w}{max(abs(w))}$          \\\hline     
               &               &sin      & $sin(w)$ &Log1p  &  $log(1+Norm(w))$              \\                                
    \hline
  \end{tabular}
    \caption{Set of operators and terminals used by the tree-based GP algorithm. Binary and unary operators are represented as B. and U., respectively.}
  \label{tab:GPParts}
\end{table}

\subsection{Fitness function}

  A critical component of a GP implementation is the definition of the fitness function. We implement the fitness evaluation as follows: At the time of evaluating a candidate program, it is applied to a training set of questions. For each question, the word vectors of the first three words are first obtained from the CBOX models. The program is  then evaluated using as arguments these three word vectors, and the program's output word vector is used to compute the quality of the program for the question.

  Let us consider the program $add(sin(ARG0),sub(ARG2,ARG1))$ and the first question in Table~\ref{tab:QUESTION_EXAMPLES} as an example.  First, we obtained the words vectors $\overrightarrow{boy}$, $\overrightarrow{girl}$, and $\overrightarrow{sons}$. Then, from the execution of the GP program, we obtain a word vector $ANS = add(sin(\overrightarrow{boy})$, $sub(\overrightarrow{sons},\overrightarrow{girl}))$. The vector $ANS$ is then presented to the CBOX model which outputs the closest word in the model.  If this word coincides with the answer to the question, a correctly answered questions counter is increased. The final fitness value is the proportion of questions in the training set that were correctly answered.

   The fitness function  serves as a direct assessment of the program quality because we can directly test whether the program produces vectors whose semantics is the one encoded by the question. However, it has an important drawback. The computational cost of repeatedly interrogating the model to determine the closest word to a given vector is very high, and it would increase with the size of the vocabulary if larger corpora were used. To diminish this cost we introduced three changes to the GP scheme.

 \begin{enumerate}
  \item Restricted vocabulary size for interrogation: The \texttt{word2vec} implementation allows the restriction of the search for the most likely word given a vector to the $l$ most frequent words in the vocabulary. Out of the total number of words ($71291$) in the vocabulary, we set $l=30000$. This reduces the computational time of the fitness function.
  \item Partial evaluation: Each program is trained on a fraction $c$ of the questions from the training set. We set this fraction to be $\frac{1}{5}$ of the size of the training set. When evaluating a program, first a subset of the questions from the training set is randomly selected, and the accuracy of the program is measured in this subset. This means that different programs are evaluated on distinct subsets of questions. 
  \item Early halt: While sequentially evaluating the questions in the (random subset of the) training set, the program does not complete the evaluation of all questions and halts  if: 1) A $NAN$ output is generated for any of the questions. 2) If after at least ten questions have been ``answered'' the proportion of correctly answered questions is at some point below $0.05$. In this case it is clearly a poorly performing program.
 \end{enumerate}
 
All the previous enhancements considerably increase the efficiency of the algorithm. While partial evaluation adds some variability in the fitness output of the programs, good programs are in general good across subsets of questions and poor programs can not specialize in niches of questions since the subset selection in the training set is made randomly. 

\section{Experiments} \label{sec:EXPE}

 The main objective of the experiments is to determine the quality of the programs generated by GP. We will compare their results for the  \emph{word analogy task} with those obtained by the application of the linear algebraic rule $\vec{d}=\vec{c}-\vec{a}+\vec{b}$, which is the one commonly used for the composition of words for this problem. In addition, we will evaluate the transferability of the best programs by applying them to a vector space  comprising  $3 \cdot 10^{6}$ vector, roughly $100$ times the size of the vector space we used to learn the programs. 

  For each fitness function and each group of questions of those described in Table~\ref{tab:QUESTION_TASK}, $30$ independent runs of the algorithms were executed. In total, $9 \times 30 = 270$ executions were conducted. Each group of questions was split into a training and test set with same number of questions. The questions in the test set were not used at any time of the evolution. 
  
\subsection{Numerical results}

 We evaluate the performance of the GP algorithms by looking at the accuracy of the best GP programs found. The accuracy, for each group of questions, is the proportion of questions correctly answered by a GP program.  For each of the $30$ runs we keep all the solutions in the last selected population ($100$ solutions by run). Among the $100$ programs, the one that has the highest accuracy in the training set is selected. Then we compute the accuracy of this program also in the test set. Using the $30$ programs, the maximum and mean accuracy are  calculated in the training and test sets.  Table~\ref{tab:RESULTS_V8} shows these values for the $9$ groups of questions. The table also shows the accuracy produced by the algebraic rule. It can be seen that the best GP evolved programs outperform the algebraic rule on all the groups of questions, although the difference in the results is more noticeable for some groups of questions (e.g., group $6$). The mean accuracy of the programs on the test set is also higher than that achieved using the algebraic rule for $6$ of the $9$ groups of questions. Notice, that since our selection of the best programs was based on the accuracy for the training set, there might be programs with a higher accuracy on the test set. We did identify some of these programs.   Interestingly, for some groups of questions (e.g., group 11) the maximum and mean accuracy in the training set is smaller than in the test set.

\begin{table}[tb]
  \centering
  \begin{tabular}{|c||c|c||c|c||c|}
    \hline
   Group     & \multicolumn{2}{|c||}{Training set} & \multicolumn{2}{|c||}{Test set} &        rule            \\ \hline
  & max.     & mean        &   max.     &   mean  &     \\         \hline  \hline
  4   &  84.21   &  82.74   &  83.66   &  82.77   &  77.70  \\\hline 
  5   &  24.67   &  21.62   &  26.46   &  21.31   &  16.16  \\\hline 
  6   &  49.34   &  41.86   &  50.98   &  41.11   &  24.92  \\\hline 
  7   &  67.09   &  66.00   &  61.27   &  58.62   &  60.44  \\\hline 
  8   &  46.83   &  44.96   &  41.11   &  38.72   &  40.40  \\\hline 
  9   &  48.08   &  45.09   &  44.56   &  39.89   &  36.83  \\\hline 
  11   &  45.71   &  41.97   &  46.70   &  44.47   &  37.94  \\\hline 
  12   &  76.16   &  73.13   &  72.38   &  69.41   &  66.50  \\\hline 
  13   &  49.69   &  40.42   &  38.77   &  32.38   &  34.21  \\ 
   \hline
  \end{tabular}
    \caption{Results of the GP algorithm in terms of the maximum and mean accuracy of the best program. Training and test sets of questions are used to evaluate the accuracy of the best program in each of the $30$ runs. The last column corresponds to the algebraic rule $\vec{d}=\vec{c}-\vec{a}+\vec{b}$.}
  \label{tab:RESULTS_V8}
\end{table}


  In a second phase of the experiments, the best program generated in the last generation of  each execution of the GP algorithm  was selected based on the sum of the training and test set accuracy values for the corresponding group of questions. We evaluated this set of $270$  programs using the  \texttt{word2vec} GoogleNews vectors. These vectors have a larger dimension ($300$ versus $200$ in the \texttt{text8} vector space) and comprise around $3$ million words. As a consequence, these vectors contain all words for the $13$ original groups of questions introduced in \cite{Mikolov_et_al:2013a}. We must remember that the reason why we did not use four of the original groups of questions was that the \texttt{text8}  vector space did not include the vector representations for all constituent words of each question in these groups. 

 Using the \texttt{word2vec} GoogleNews vectors, we can test the evolved programs in all the data sets. The same set of operations encoded in the programs are applied, but this time using the new vector representation. The output vector is then submitted to the model that determines whether the closest word in the space of  \texttt{word2vec} GoogleNews vectors is the right answer to the question.

The results of this evaluation are shown in Table~ \ref{tab:RESULTS_F0}, where each row corresponds to one of the $13$ original groups of questions. Each column $j$ shows the best accuracy produced by the best program among the $30$ generated with function $F0$ for the group of questions represented in column $j$. The last column shows the accuracy results of the algebraic rule. In each row, all the programs that produce results better than the one in the last column are shown in bold. 

 Notice that we can evaluate the $270$ GP program in all $13$ groups of questions independently of the group used to learn them. Since all the questions have the same structure, we can apply the programs to them. 

\begin{table*}[tb]
  \centering
  \begin{tabular}{|c||c|c|c|c|c|c|c|c|c||c|}
    \hline
  Group  & 4     & 5        &   6      &   7      &   8      &   9      &  11      &   12     &  13      &  rule   \\         \hline  \hline
  1   &  77.80   &  63.41   &  51.43   &  {\bf{81.72}}   &  81.49   &  76.23   &  81.41   &  78.60   &  81.49   &  81.49  \\\hline 
  2   &  14.68   &  10.87   &  2.08   &  {\bf{16.18}}   &  15.95   &  14.68   &  {\bf{16.07}}   &  {\bf{16.07}}   &  15.95   &  15.95  \\\hline 
  3   &  65.94   &  41.16   &  33.62   &  72.75   &  72.75   &  69.06   &  68.98   &  70.48   &  {\bf{74.17}}   &  72.75  \\\hline 
  4   &  73.27   &  62.18   &  48.71   &  77.62   &  77.62   &  73.66   &  77.23   &  77.62   &  77.62   &  77.62  \\\hline 
  5   &  27.75   &  18.77   &  17.66   &  28.86   &  28.86   &  21.39   &  {\bf{29.87}}   &  28.86   &  28.86   &  28.86  \\\hline 
  6   &  {\bf{34.16}}   &  27.87   &  28.61   &  {\bf{32.68}}   &  32.43   & {\bf{34.28}}   &  {\bf{33.54}}   &  32.31   &  32.43   &  32.43  \\\hline 
  7   &  88.66   &  63.19   &  71.83   &  89.56   &  89.56   &  86.33   &  {\bf{89.93}}   &  89.26   &  89.56   &  89.56  \\\hline 
  8   &  65.21   &  48.26   &  33.27   &  67.53   &  67.53   &  46.65   &  {\bf{69.05}}   &  {\bf{68.06}}   &  67.53   &  67.53  \\\hline 
  9   &  70.90   &  60.00   &  61.90   &  73.18   &  73.18   &  61.99   &  71.09   &  72.70   &{\bf{74.41}}   &  73.18  \\\hline 
  10   &  85.17   &  83.79   &  71.78   &  85.42   &  85.42   &  82.23   &  85.36   &  85.36   &  85.42   &  85.42  \\\hline 
  11   &  62.92   &  43.30   &  48.69   &  {\bf{68.44}}   &  66.20   &  64.27   &  64.46   &  66.13   &  66.20   &  66.20  \\\hline 
  12   &  76.93   &  74.76   &  74.68   &  78.36   &  78.36   &  75.13   &  78.21   &  78.14   &  78.36   &  78.36  \\\hline 
  13   &  55.12   &  21.75   &  29.69   &  {\bf{65.71}}   &  63.87   &  56.85   &  62.83   &  62.83   &  63.87   &  63.87   \\ \hline  
  \end{tabular}
    \caption{Results of the best programs produced by the GP algorithm on all groups of questions when  the set of GoogleNews vectors are used to execute the GP programs. Last column corresponds to the algebraic rule $\vec{d}=\vec{c}-\vec{a}+\vec{b}$.}
  \label{tab:RESULTS_F0}
\end{table*}

\begin{table*}[tb]
  \centering
  \begin{tabular}{|c||c|c|c|c|c|c|c|c|c|}
    \hline
  Group  & 4     & 5        &   6      &   7      &   8      &   9      &  11      &   12     &  13       \\         \hline  \hline
  1   &  23   &  52   &  76   &  104   &  129   &  153   &  185   &  235   &  261  \\\hline 
  2   &  9   &  52   &  76   &  104   &  129   &  153   &  185   &  235   &  261  \\\hline 
  3   &  15   &  58   &  76   &  120   &  129   &  153   &  198   &  235   &  247  \\\hline 
  4   &  23   &  58   &  76   &  120   &  129   &  178   &  185   &  235   &  261  \\\hline 
  5   &  23   &  52   &  76   &  120   &  129   &  178   &  208   &  235   &  261  \\\hline 
  6   &  23   &  58   &  76   &  111   &  129   &  178   &  196   &  224   &  261  \\\hline 
  7   &  23   &  58   &  76   &  120   &  129   &  178   &  208   &  235   &  261  \\\hline 
  8   &  23   &  52   &  76   &  120   &  129   &  178   &  208   &  235   &  261  \\\hline 
  9   &  23   &  58   &  76   &  104   &  129   &  153   &  195   &  235   &  247  \\\hline 
  10   &  23   &  58   &  76   &  120   &  129   &  178   &  208   &  224   &  261  \\\hline 
  11   &  23   &  58   &  76   &  104   &  129   &  178   &  208   &  235   &  261  \\\hline 
  12   &  23   &  37   &  76   &  120   &  129   &  154   &  208   &  235   &  261  \\\hline 
  13   &  23   &  58   &  76   &  104   &  129   &  153   &  195   &  235   &  261  \\ \hline
  \end{tabular}
    \caption{Indices of the best programs produced by the GP algorithm on all groups of questions.}
  \label{tab:RESULTS_BEST_PROG}
\end{table*}

  There are a number of remarkable facts in the results shown in Table~\ref{tab:RESULTS_F0}: 

 \begin{enumerate}
  \item Some of the programs improved the accuracy for groups of questions that were  not  in the original reduced benchmark of $9$ groups. This is the case for the group of questions $3$.
  \item The best program for the group of questions $j$ is not, in general, a program evolved to answer this group of questions. 
  \item There are programs evolved for some groups of questions that are good at answering questions for all groups. For example, this happens with programs learned using the group of questions $12$. 
 \end{enumerate}

\subsection{Evaluating answers and evolved programs}

  One important issue is the interpretability of the evolved programs and how are they related with the algebraic rule. Out of the $270$ programs tested, $8$ were equivalent to the algebraic rule. Four of these programs are shown in Figure~\ref{fig:EQUIV_TREES}. It can be seen how the same rule is implemented in distinct ways using only the operators $add$, $sub$, and $neg$. These results show that, as an algorithm to create word compositions, GP can automatically learn compositional methods designed by humans. 
 
\begin{figure}[htb]
   \begin{center}    
\includegraphics[width=4.0cm]{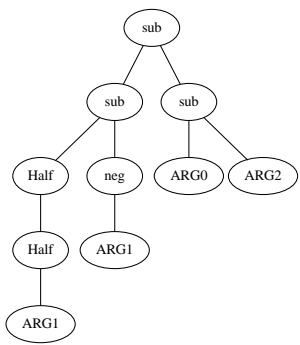}
\caption{Program number $247$, according to the indices in Table~\ref{tab:RESULTS_BEST_PROG}. It was learned from the group of questions $13$ and produced the best accuracy, among the $270$ programs selected, for group of questions $3$. Its accuracy  using the  \texttt{word2vec} GoogleNews vectors was $74.17$, above the $72.75$ accuracy of the algebraic rule for the same group of questions. See tables~\ref{tab:RESULTS_F0} and~\ref{tab:RESULTS_BEST_PROG} for details of the program behavior.}
\label{fig:GOODTREE_247}
\end{center}    
\end{figure}

We also analyzed those GP programs that outperformed the algebraic rule. An exemplar of this type of programs is shown in Figure~\ref{fig:GOODTREE_247}. It was the best program found for the group of questions $3$. Its accuracy  using the  \texttt{word2vec} GoogleNews vectors was $74.17$, above the $72.75$ accuracy of the algebraic rule for the same group of questions. 

\begin{figure}[htb]
   \begin{center}    
     \includegraphics[width=4.0cm]{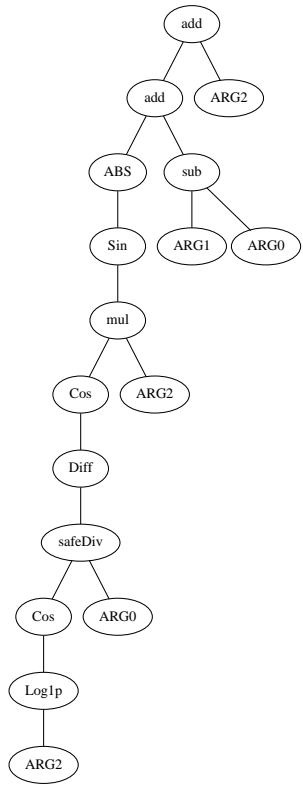}
   \caption{Program $104$, learned from the group of questions $7$. See tables~\ref{tab:RESULTS_F0} and~\ref{tab:RESULTS_BEST_PROG} for details of the program behavior.}
\label{fig:GOODTREE_104}
 \end{center}    
\end{figure}

\begin{figure}[htb]
   \begin{center}    
\includegraphics[width=4.0cm]{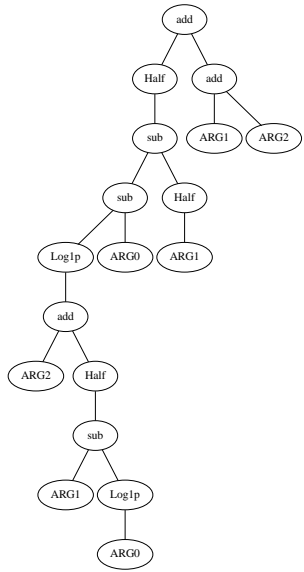}
\caption{Program $178$, learned from the group of questions $9$.}
\label{fig:GOODTREE_178}
\end{center}    
\end{figure}

\begin{figure}[htb]
   \begin{center}    
     \includegraphics[width=6.0cm]{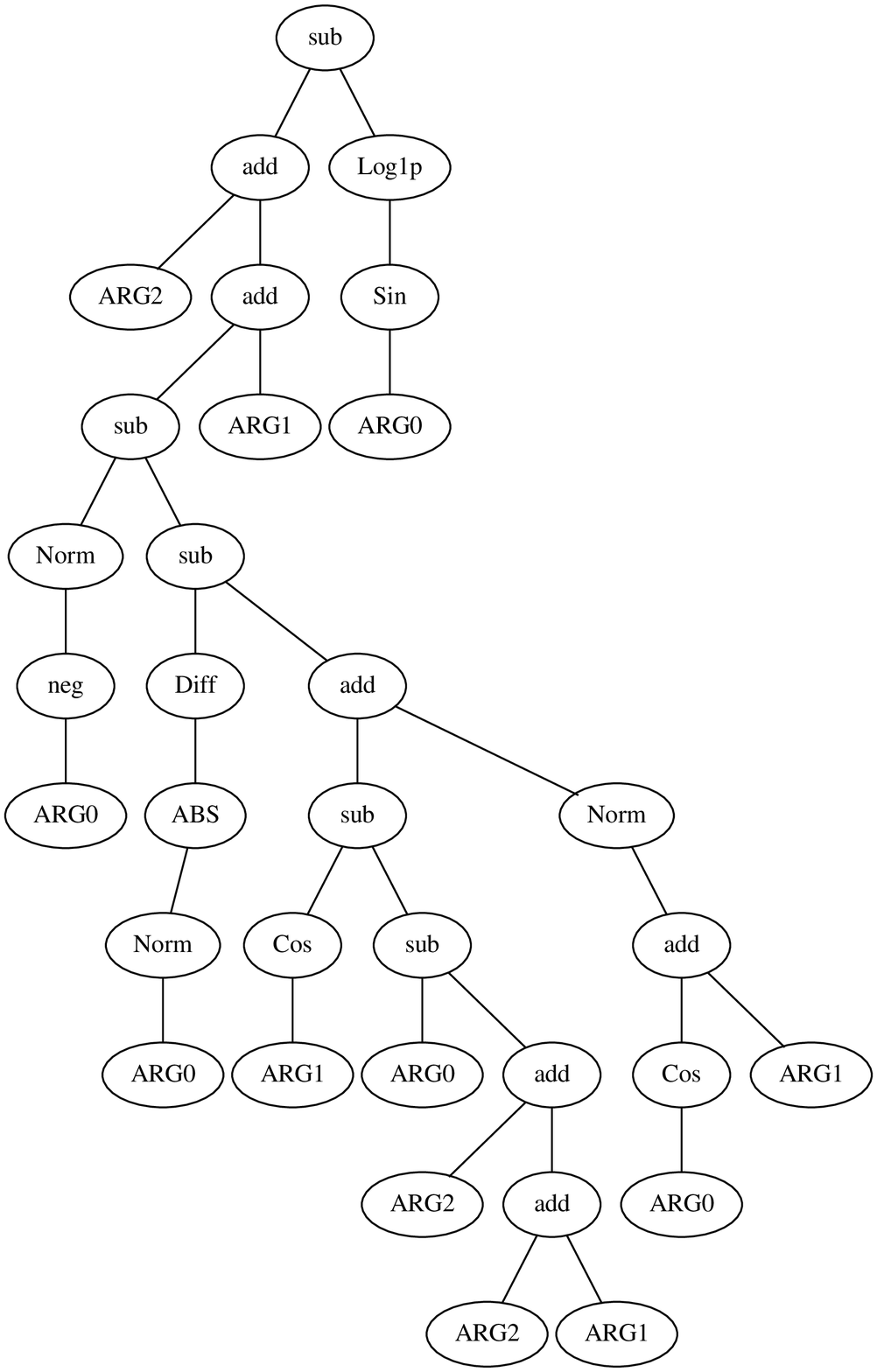}
  \caption{Program $185$, learned from the group of questions $11$.}
\end{center}    
\label{fig:GOODTREE_185}
\end{figure}

\begin{figure}[htb]
   \begin{center}    
\includegraphics[width=4.0cm]{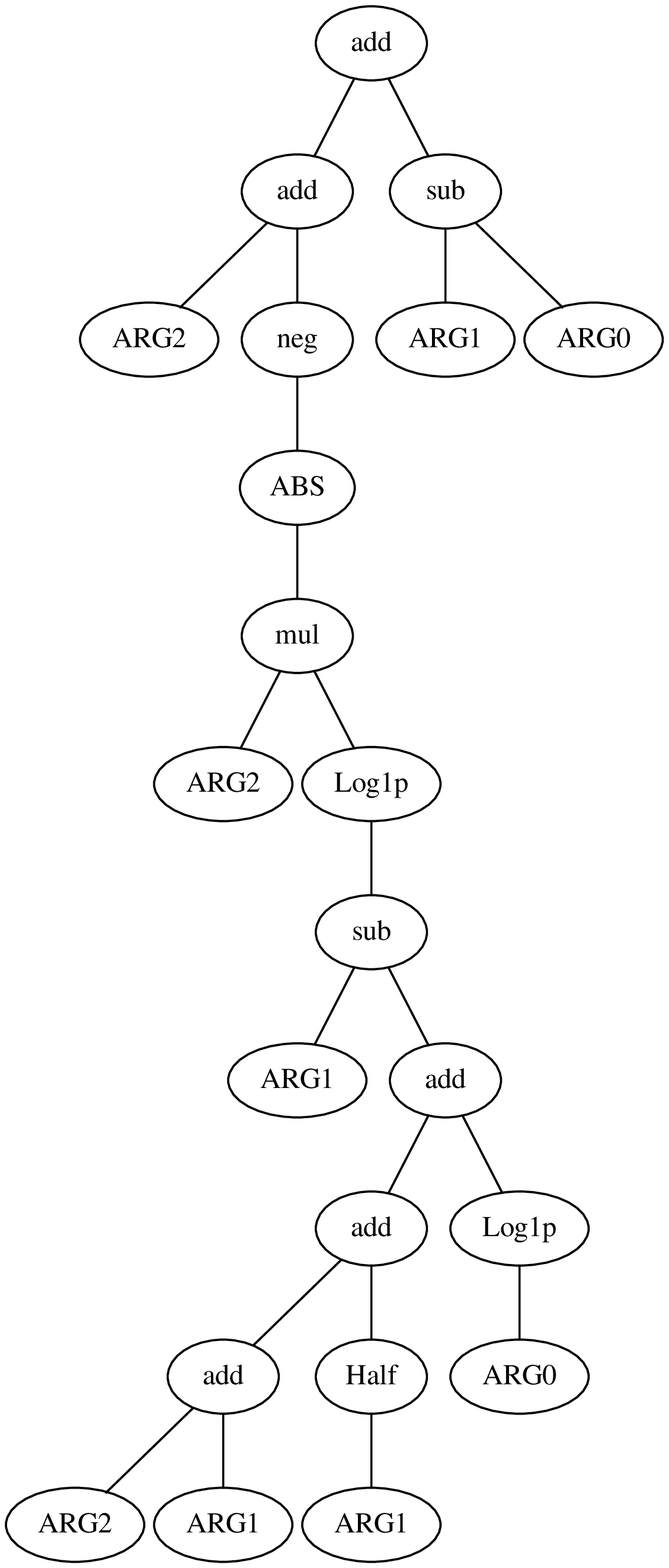}
   \caption{Program $208$, learned from the group of questions $11$.}
\label{fig:GOODTREE_208}
\end{center}    
\end{figure}

\begin{figure}[htb]
   \begin{center}    
     \includegraphics[width=4.0cm]{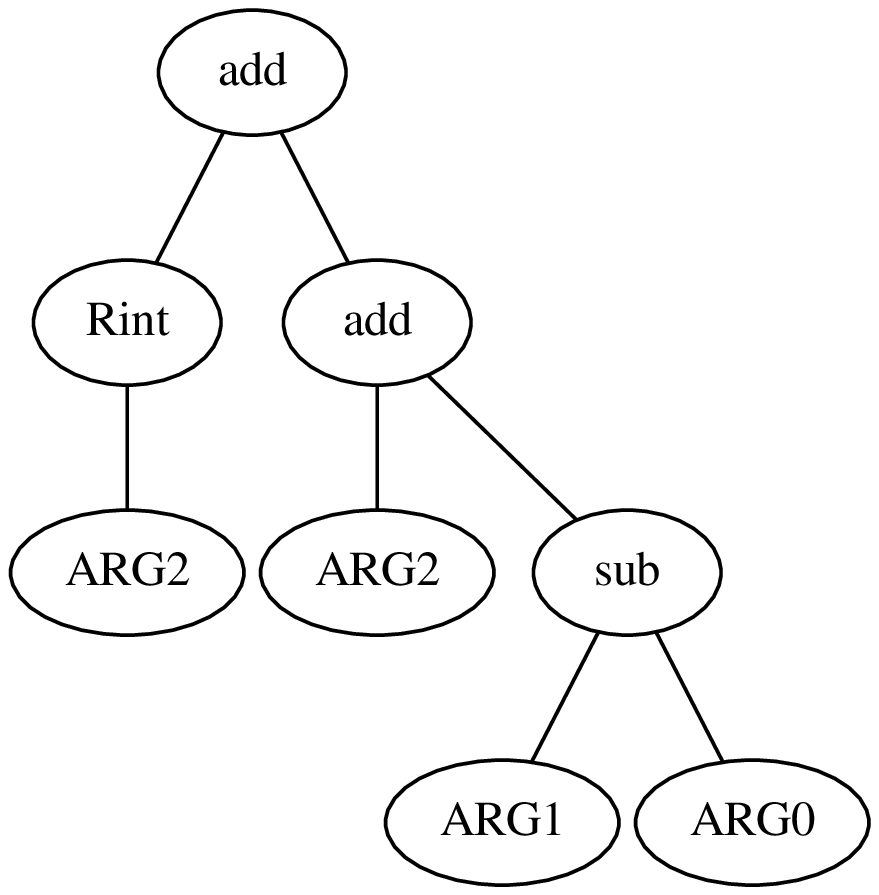}
       \caption{Program $235$, learned from the group of questions $12$.}
\label{fig:GOODTREE_235}
 \end{center}    
\end{figure}

 The tree shown in Figure~\ref{fig:GOODTREE_247} is a slight modification of the algebraic rule. Instead of adding $ARG2$ to the rule, this programs adds $\frac{5}{4}ARG2$ and this change allows it to increase the accuracy for the group of questions. A trend observed in other evolved programs was that they contained building blocks from the algebraic rule. As in the case of the programs shown in Figure~\ref{fig:EQUIV_TREES}, these structural features were not specifically induced, they were acquired as part of the evolutionary process. Other programs that produced high accuracy values are shown in figures~\ref{fig:GOODTREE_104}-~\ref{fig:GOODTREE_235}. When analyzing the behavior of these programs, tables~\ref{tab:RESULTS_F0} and~\ref{tab:RESULTS_BEST_PROG} should be consulted.

 \subsection{Discussion}
 
 We go back to the research questions posed at the beginning of this work and try to answer them based on the results of our experiments. 

\begin{itemize}
  \item  For embedding representations, can meaningful vector algebraic operations be learned  from training examples? Yes, they can be learned within a relatively small computational time. 
  \item  If so, is GP a feasible approach to do it?  Yes, GP is a natural solution for this type of problem and even straightforward implementations can deal with the problem. 
  \item  How do GP programs score with regard to the algebraic rule commonly applied on vector representations? GP programs can learn the same rule designed by humans and, therefore, can reach the same results. They can also outperform these results but, at least for the  class of word vector representation and the basic tree-based GP approach implemented, the improvements are moderate. 
  \item Are GP evolved programs transferable across linguistic tasks, vector representations and corpora? Definitively. The high transferability of the programs across groups of questions may be supported by the general underlying commonality between analogies that these group of questions represent. However, it is remarkable how the programs can be transferred to a vector space where both the dimension of the vectors, and the number of vectors increase dramatically. In this respect, transferability opens an  additional opportunity for efficiency gain. Programs can be learned using small vector spaces, and then validated or refined on more computationally costly large vector spaces. 
\end{itemize}

\section{Conclusions and future work} \label{sec:CONCLU}

While semantic spaces and word vector representations are able to capture some of the semantic relationships between the words, compositional methods are necessary to extend their use to multi-word constructions. In this paper we have proposed representing compositional vector operations as simple programs that can be automatically learned from data.  We have shown that, using GP, it is possible to encode a set of vector operations as a program, that the programs can be evolved to achieve higher accuracy than the human rules conceived to manipulate the words, and that the programs are valid for datasets other than those from which they have been learned, i.e., they are transferable programs. Furthermore, our results indicate that it is possible to learn programs using vector vocabularies of small to moderate sizes and then test them in bigger domains where the evaluation of a program is more costly.

\subsection{Future work}

As lines for future work we consider the following:

 \subsubsection{Use alternative methods for the word vector generation}

  While GP approaches can explore a vast range of possible word compositions, the usefulness of more intricate programs is, to some extent, constrained by the nature of the relationships that the vectors can encode. For example, if the methods used to construct the embeddings do not allow  non-linear relationships between the vectors, then the improvements of the GP programs over plain linear algebraic compositional operators will be marginal. Therefore, it would be important to test the automatic generation of word compositions with GP on word vectors generated using diverse methods. 

 \subsubsection{Evolve functions for the similarity metric}

 Since it has been shown that the type of similarity metric can critically influence the accuracy results \cite{Levy_and_Goldberg:2014}, it makes sense to learn this function as well. One difficulty is that the output of this function will be a numerical value and not a vector like the other operators used in the current GP representation. In addition,  evaluating an alternative similarity metric implies using the candidate metric to compute distances to all vectors in the vector space, a process that can be very costly computationally. 

 \subsubsection{Combining different word representations}

  Turian et al \cite{Turian_et_al:2010} have shown that combining different word representations can improve accuracy for supervised classification tasks in NLP \cite{Turian_et_al:2010}. We envision the evolution of programs which are able to combine different word vector representations.
 
  \subsubsection{Using more sophisticated GP approaches}

   From the point of view of research in genetic programming, word embeddings open an interesting research line. More research is needed to identify which, among more sophisticated GP approaches, are the most appropriate for their application to semantic spaces. Among possible lines of research are the following: 

\begin{itemize} 
  \item Alternative GP representations: In addition to trees, other GP representations such as grammars  \cite{Oneil_and_Ryan:2003,Bosman_and_DeJong:2004} and Cartesian GP \cite{Miller_and_Thomson:2000} could be considered.
 \item More complex descriptions of the compositional operators: One open question is to what extent can more complex functions better exploit the underlying semantic relationships between the word vectors. This could be investigated by adding other algebraic operators to the set of GP functions, including ternary operators. Another possibility is representing the composition of vectors with  ensembles of GP programs  \cite{Bhowan_et_al:2013}. 
  \item Reusing problem information: Approaches able to identify and transfer building blocks \cite{Iqbal_et_al:2014} between word vectors or corpora of varying dimensions  arise as potential candidates. 
 \item Behavioral program synthesis:  One direction in which the evolution of the programs could be improved is by analyzing and assessing the quality of  the intermediate vectors produced in the evaluation of  the programs. In general, algorithms that advocate a more efficient use of the information displayed by the behavior of the GP programs \cite{Krawiec_et_al:2016} could lead to better solutions  and reveal additional insights in learning compositional methods.
\end{itemize}

\subsubsection*{Acknowledgments}

This work has received support through through the IT-609-13 program (Basque Government), TIN2016-78365-R (Spanish Ministry of Economy, Industry and Competitiveness) and Brazilian CNPq Program Science Without Borders No.: 400125/2014-5.

\bibliographystyle{abbrv}

\end{document}